%% file: main.tex
\documentclass[conference]{IEEEtran}
\IEEEoverridecommandlockouts
\usepackage{cite}
\usepackage{amsmath,amssymb,amsfonts}
\usepackage{algorithmic}
\usepackage{graphicx}
\usepackage{textcomp}
\usepackage{xcolor}
\usepackage{tikz}
\usepackage[most]{tcolorbox}
\usepackage{hyperref}

\def\BibTeX{{\rm B\kern-.05em{\sc i\kern-.025em b}\kern-.08em
    T\kern-.1667em\lower.7ex\hbox{E}\kern-.125emX}}
\begin{document}

\title{Leveraging Large Language Models for Solving Rare MIP Challenges\\
% {\footnotesize \textsuperscript{*}Note: Sub-titles are not captured for https://ieeexplore.ieee.org  and
}

% \and
% \and
% \and
% \and
% \and

\author{\IEEEauthorblockN{1\textsuperscript{st} Teng Wang}
\IEEEauthorblockA{\textit{Department of Mathematics} \\
\textit{University of Hong Kong}\\
Hong Kong, China \\
wt0318@connect.hku.hk}
\and
\IEEEauthorblockN{2\textsuperscript{nd} Wing-Yin Yu}
\IEEEauthorblockA{\textit{Noah’s Ark Lab} \\
\textit{Huawei}\\
Hong Kong, China \\
rocket.YuWingYin@huawei.com}
\and
\IEEEauthorblockN{3\textsuperscript{rd} Ruifeng She}
\IEEEauthorblockA{\textit{Noah’s Ark Lab} \\
\textit{Huawei}\\
Hong Kong, China \\
sheruifeng@huawei.com}
\and
\IEEEauthorblockN{4\textsuperscript{th} Wenhan Yang}
\IEEEauthorblockA{\textit{Department of Mathematics} \\
\textit{University of Hong Kong}\\
Hong Kong, China \\
u3621353@connect.hku.hk}
\and
\IEEEauthorblockN{5\textsuperscript{th} Taijie Chen}
\IEEEauthorblockA{\textit{Department of Civil Engineering} \\
\textit{University of Hong Kong}\\
Hong Kong, China \\
ctj21@connect.hku.hk}
\and
\IEEEauthorblockN{6\textsuperscript{th} Jianping Zhang }
\IEEEauthorblockA{\textit{Noah’s Ark Lab} \\
\textit{Huawei}\\
Hong Kong, China \\
zhang.jianping4@huawei.com}

}

\maketitle

\input{sections/0abstract}

\input{sections/1introduction}

\input{sections/2related_work}

\input{sections/new_method}

\input{sections/5experiment}

\input{sections/6conclusion}

\input{sections/8acknowledge}
\bibliographystyle{IEEEbib}
\bibliography{main}

\end{document}

%% file: sections/0abstract.tex
\begin{abstract}
Mixed Integer Programming (MIP) has been extensively applied to areas requiring mathematical solvers to address complex instances within tight time constraints. However, as the problem scale increases, the complexity of model formulation and finding feasible solutions escalates significantly. Beneficial from outstanding text generation capacity of Large Language Models (LLMs), building and solving industrial-level instances becomes insensitive to problem scale. While LLMs, like GPT-4, can handle some traditional medium-scale MIP problems, they struggle with uncommon or highly specialized MIP scenarios. Fine-tuning LLMs can yield some feasible solutions for medium-scale MIP instances, but these models typically fail to explore diverse solutions when constrained by a low and constant temperature. In this paper, we propose and evaluate a recursively dynamic temperature method integrated with a chain-of-thought approach to exploit a large feasible region. Our findings show that starting with a high temperature and gradually lowering it leads to better feasible solutions compared to other dynamic temperature strategies. Additionally, by comparing results generated by the LLM with those from Gurobi, we demonstrate that the LLM can produce solutions that complement traditional solvers by accelerating the pruning process and improving overall efficiency.

\end{abstract}
\begin{IEEEkeywords}
Mixed integer programming, Chain-of-Thought (CoT), dynamic temperature, ride pooling, bipartite matching.
\end{IEEEkeywords}

%% file: sections/1introduction.tex
\section{Introduction}

\label{sec:intro}

Mixed Integer Programming (MIP) is a fundamental tool in many optimization domains, such as the Traveling Salesman Problem (TSP) \cite{laporte1992traveling} and facility location planning \cite{klose2005facility}. MIP also plays a particularly critical role in time-sensitive applications like transportation and network scheduling \cite{he2018vehicle}, where finding a feasible solution within a short time frame is essential to maintaining system operability and avoiding downtime.

The traditional approach to solve MIP problems is the branch-and-bound (B\&B) algorithm\cite{lawler1966branch}. While this method guarantees to find the optimal solution for a given instance, the efficiency of mathematical solvers that use such method like Gurobi\cite{achterberg2019s} diminishes as the problem scale increases \cite{jablonsky2015benchmarks}. Moreover, the complexity of model formulation grows significantly with the dimensionality of the problem. For example, the growth rate of the complexity of a 3D bin-packing problem \cite{martello2000three} is considerably higher than that of a 2D bin-packing problem \cite{johnson1974fast}.

To expedite the search for optimal solutions, mathematical solvers implement various techniques such as heuristics, cutting planes, parallelism, presolve\cite{gomory2010outline}. However, despite these advanced methods, solvers still face challenges in efficiently handling large-scale MIP problems within tight time constraints. Large language models (LLMs), with their strong pattern recognition capabilities, can achieve similar objectives with only minimal data and modeling information. For instance, Yang et al. \cite{yang2023large} pioneered the application of Chain-of-Thought (CoT) reasoning \cite{wei2022chain} in large language models such as GPT-3.5 \cite{brown2020language} and GPT-4 \cite{achiam2023gpt} to address problems like the TSP using only the coordinates of cities, without explicitly requiring distances between each pair of cities. This approach reduces the time complexity from $O(n^2)$, typically required for distance calculations in traditional mathematical solvers, offering a more efficient solution.

However, the previous work by Yang et al. \cite{yang2023large} has several drawbacks. First, the instance data is generated from randomly sampled integers, which may reduce its validity as a demonstration of the LLMs capabilities in real-world MIP applications. Second, the TSP is a well-known and extensively studied problem, meaning LLMs have been trained on similar data and the same TSP model numerous times. In our experiments, we observed that LLMs often struggle with complicated mathematic models and frequently fail to grasp the MIP modeling process. These factors raise concerns about the robustness of LLMs in real-world applications. 

Our work focuses on how to integrate LLMs into real-world applications. To demonstrate the generalizability of LLMs in real-world scenarios, we developed a fine-grained simulator and utilized the operational dataset provided by DiDi in November 2016\cite{yao2018deep} to simulate the passenger-driver matching process in the ride-pooling market using MIP. Our work is divided into three main components: (1) We construct a carpooling MIP model based on real-world data while capturing and storing vehicle locations, order locations, MIP instances, and intermediate feasible solution statuses for future training purposes. (2) Leveraging the pattern recognition capabilities of LLMs and CoT reasoning, we generate prompts using only abstract information from the carpooling dataset, bypassing the need to compute MIP parameters, like the distance between the vehicle and the user's order, explicitly. We then perform supervised fine-tuning on LLaMA 3.1 (8B) \cite{dubey2024llama} to discover better feasible solutions, comparing these with the top three feasible solutions generated by traditional mathematical solvers. The results from LLM can be used to accelerate the pruning process in conventional mathematical solvers. (3) We employ recursive dynamic temperature adjustments to refine the quality of feasible solutions generated by the LLM. Through a strategy of starting at a higher temperature and gradually reducing it, we observe significant improvements in solution quality. By systematically evaluating performance under various temperature schedules, we identify the highly effective strategy for enhancing the effectiveness and consistency of the solutions produced.

%% file: sections/2related_work.tex
\section{Related Work}
\label{sec:relatedwork}
MIP plays a crucial role in combinatorial optimization, with applications in planning\cite{klose2005facility}, scheduling\cite{xiong2022survey}, and routing\cite{braekers2016vehicle}. Traditional methods like B\&B \cite{lawler1966branch} have been widely used to solve MIP problems. However, these methods can be computationally intensive, leading to growing interest in enhancing MIP solvers with machine learning (ML) and LLMs.

Recent works integrating ML with MIP can be categorized into two main approaches \cite{zhang2023survey}: exact algorithms and heuristic algorithms. For exact methods like B\&B, ML models have been used to optimize branching variable selection and node selection, significantly improving solution efficiency \cite{gasse2019exact} \cite{khalil2016learning}. On the heuristic side, techniques like Large Neighborhood Search and Feasibility Pump have benefited from ML integration, leading to higher solution quality and computational efficiency \cite{song2020general} \cite{qi2021smart}. Additionally, Graph Neural Networks have been leveraged to represent MIP instances, enhancing decision-making processes like branching and node selection \cite{gasse2019exact}. Reinforcement learning is also increasingly applied in both exact and heuristic methods to support adaptive decision-making within the B\&B framework \cite{tang2020reinforcement}.
\input{tables/building_model_time}

With the advent of LLMs and the rise of AI agents, more research has focused on translating natural language into operations research problems \cite{xiao2023chain, ahmaditeshnizi2023optimus, MLPrompt}. Although zero-shot learning typically performs poorly on complex problems, LLMs have significant potential, and their performance can be improved through techniques like the chain of thought \cite{wei2022chain}, tree of thought \cite{yao2023tree}, and self-consistency \cite{wang2022self}. Yang et al.’s work \cite{yang2023large} utilizes models like PaLM \cite{chowdhery2023palm} and GPT-4 \cite{achiam2023gpt} to tackle linear regression and the TSP with CoT reasoning, demonstrating success on small-scale problems. Our work fine-tunes the LLaMA 3.1 (8B) model \cite{dubey2024llama} using both model information and real MIP instance data capable of generating feasible solutions, and proposes an adaptive temperature strategy that iteratively enhances LLM performance, leading to the generation of even more optimized feasible solutions.

%% file: tables/building_model_time.tex
\begin{figure}[h!]
    \centering
    \includegraphics[width=0.4\textwidth]{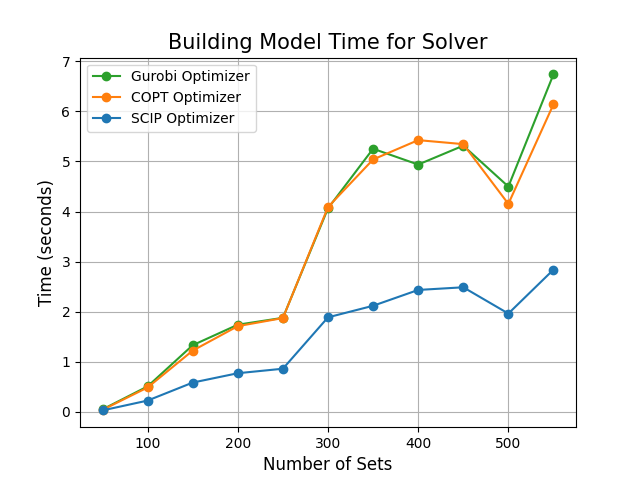} 
  \caption{The time cost of building the model increases significantly as the problem scale grows. This trend illustrates the growing computational complexity associated with larger problem instances.}

    \label{imgs:buildingModelTime}
\end{figure}

%% file: sections/new_method.tex
\section{Method}
\label{sec:method}

Given that LLMs have been trained on numerous traditional MIP problems, such as the TSP, and considering the limitations in the generalizability of previous work due to the use of non-real-world data, we aim to assess the potential of LLMs in MIP under real-world conditions. To achieve this, we construct a carpooling MIP model and develop a simulator to replicate the vehicle dispatching process. The DiDi operational dataset, which consists of the location and time of orders from November 2016, serves as the foundation for generating real-world data in this study. The input is a long text containing information about the locations of orders and the positions of various categories of vehicles, while the output is feasible solutions for dispatching these vehicles to different users.

\subsection{Problem Statement}
% There are two types of vehicles: (1) empty vehicles and (2) vehicles with one passenger. Additionally, we assume that each order is associated with a single customer and that a vehicle can accommodate at most two passengers at a time. Our goal is to minimize the total distance traveled for picking up customers, subject to the above constraints. The detailed method for calculating the distance in various scenarios is provided in Appendix\ref{sec:appendix:distance}.
There are two types of vehicles: (1) empty vehicles and (2) vehicles with one passenger. Additionally, we assume that each order is associated with a single customer and that a vehicle can accommodate at most two passengers at a time. Our goal is to minimize the total distance traveled for picking up customers, subject to the above constraints. Simplify the model by calculating Manhattan distances when formulating the MIP. However, in the simulation process, Dijkstra’s algorithm is employed to simulate vehicle movement.

The notation is as follows:

\noindent- \( x_{ij} \) is a decision variable indicating whether empty car \( i \) is assigned to user \( j \).

\noindent- \( y_{ijk} \) is a decision variable indicating whether empty car \( i \) is assigned to pick up user \( j \) and then user \( k \).

\noindent- \( z_{ij} \) is a decision variable indicating whether car \( i \) with one passenger willing to share is assigned to user \( j \).

\noindent- \( d_{ij} \) is the distance between vehicle \( i \) and user \( j \).

\noindent- \( d_{jk}' \) is the distance between user \( j \) and user \( k \).

\input{tables/workflow}

\noindent- \( d_{ij}'' \) is the distance between vehicle \( i \) (with one order) and user \( j \).

\noindent- \( m \) is the number of empty vehicles.

\noindent- \( n \) is the number of vehicles with one order.

\noindent- \( p \) is the number of orders.

\textbf{1. Objective Function}

The objective is to minimize the total distance between vehicles and passengers across all segments of the vehicle paths:
$$
\scalebox{0.85}{%
    \begin{minipage}{0.55\textwidth}
    \[
    \min \sum_{i=1}^{n_1} \sum_{j=1}^{m} x_{ij} \cdot d_{ij} + \sum_{i=1}^{n_1} \sum_{j=1}^{m} \sum_{k=1, k \neq j}^{m} y_{ijk} \cdot (d_{ij} + d_{jk}') + \sum_{i=1}^{n_2} \sum_{j=1}^{m} z_{ij} \cdot d_{ij}''
    \]
    \end{minipage}%
}
$$

\textbf{2. Constraints}
\label{sec:modeling:constraints}

Order Coverage: Each user is assigned to exactly one vehicle:
   \[
   \sum_i x_{ij} + \sum_i \sum_{k, j \neq k} (y_{ijk} + y_{ikj}) + \sum_{i'} z_{i',j} = 1, \ \forall \ j
   \]

Vehicle Capacity for Empty Vehicles: Each empty vehicle picks up at most two people:
   \[
   \sum_j x_{ij} + \sum_j \sum_{k \neq j} y_{ijk} \leq 1, \ \forall \ i
   \]

Vehicle Capacity for Shared Rides: Each vehicle with one passenger picks up at most one more person:
   \[
   \sum_j z_{ij} \leq 1, \ \forall \ i
   \]

Binary Variable Constraints:
   \[
   x_{ij}, y_{ijk}, z_{i,j} \in \{0, 1\}, \ \forall \ i, j, k
   \]

\textbf{3. Analysis of the Constraint Matrix}\\
As discussed in the previous constraints part, Figure \ref{imgs:buildingModelTime} illustrates the relationship between the model-building time and the number of sets. The shape of the constraint matrix is given by:

\[
(m+n+p, \; m \cdot p + m \cdot p^2 + n \cdot p)
\]
\input{tables/buildingRate}

The complexity of the constraint matrix increases significantly as the size of the set grows, resulting in a very high level of computational complexity. This makes it infeasible to solve the problem instantly if the size of the set increases significantly.

\subsection{Data Collection and Prompt Generation}

Prior to fine-tuning the LLMs, we first identify the positions of the orders and the various types of vehicles. The next step is to generate the labels, which consist of the intermediate feasible solution and the optimal solution for this MIP instance.

\subsubsection{Data Storage}
When the simulator constructs and solves the MIP instances using the Gurobi solver, we capture several critical pieces of information that include (1) the location of every vehicle and order, (2) the status of intermediate feasible solutions, and (3) the optimal solution.

\subsubsection{Prompt Generation}

% Following the collection of raw and intermediate data, we compile the LaTeX code for the model-building process, the details of each vehicle and order scenario, and both feasible and optimal solutions into a text format, as presented in Appendix \ref{sec:appendix:prompt}. This information is then used to generate the desired prompt for subsequent training and inference processes.
Following the collection of raw and intermediate data, we compile the LaTeX code for the model-building process, the details of each vehicle and order scenario, and both feasible and optimal solutions into a text format. This information is then used to generate the desired prompt for subsequent training and inference processes.

\subsection{Recursive CoT with dynamic temperature}

Since LLMs, after fine-tuning, exhibit a strong ability to grasp problem patterns, they often produce the same feasible solution at lower temperatures—even when the prompt suggests that this solution isn’t optimal. Despite recursive adjustments from lower temperatures to higher temperatures intended to explore a broader solution space, LLMs may still become trapped in the previous bad solutions. Thus, a temperature strategy is needed to effectively explore diverse possibilities and enhance solution quality.

To address this, we employ a recursive approach with dynamic temperature to leverage CoT effectively. Our strategy involves initially generating feasible solutions with a high temperature to explore a broader solution space. We then iteratively refine these solutions by gradually lowering the temperature. This dynamic temperature adjustment begins with a high temperature to facilitate exploration and progressively decreases to a low temperature to focus on refinement. This balanced approach helps us improve solution diversity and quality, ultimately leading to better feasible solutions. Figure \ref{imgs:workflow} illustrates the entire workflow of training and inference.

%% file: tables/workflow.tex
\begin{figure}[h!]
    \centering
    \includegraphics[width=0.44\textwidth]{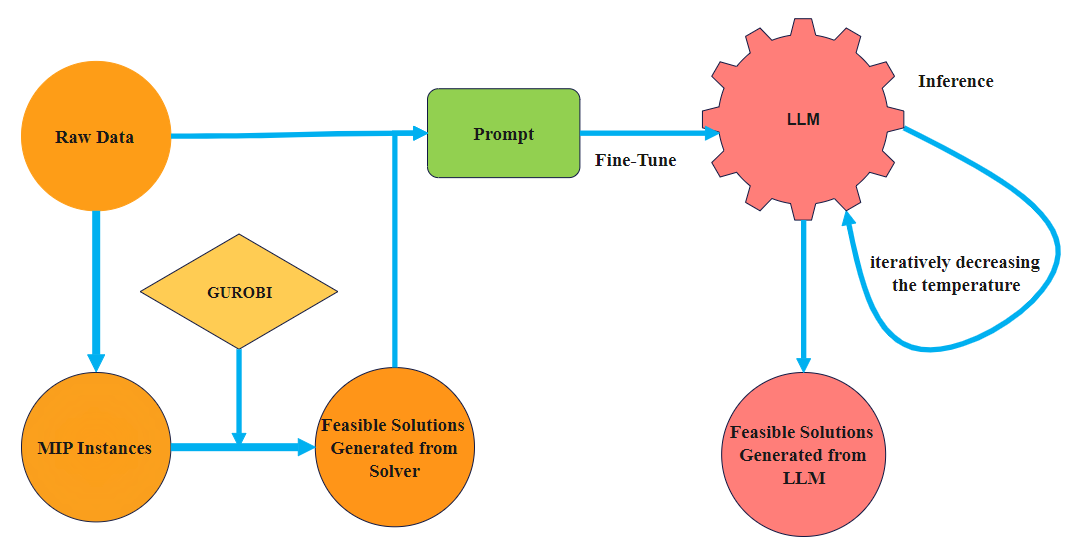} 
  \caption{Workflow of training and inference, showing the recursive approach where the temperature starts high to explore a broad solution space and gradually decreases to refine and improve solution quality.}

    \label{imgs:workflow}
\end{figure}

%% file: tables/buildingRate.tex
\begin{figure}[h!]
    \centering
    \includegraphics[width=0.40\textwidth]{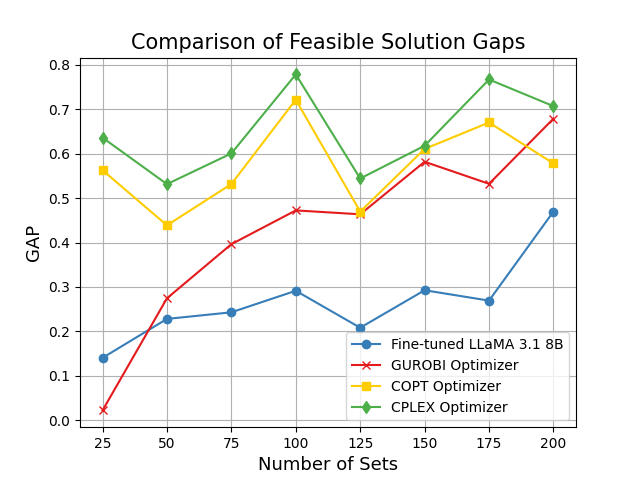} 
    \caption{The gap, defined as the difference between the feasible and optimal solution, shows that a larger gap means worse performance. After fine-tuning, LLaMA 3.1 8B generates feasible solutions with a smaller gap than solver’s first three solutions, especially as MIP instance scale grows.
}
    \label{imgs:comparisionLLMandGurobi}
\end{figure}

%% file: sections/5experiment.tex
\section{Experiment}
\label{sec:experiment}

% 等等考虑加一个图片来表示LLMs求解是线性关系，但是gurobi建模是几何倍的时间

Using LLMs to retrieve the exact optimal solution for medium and large-scale MIP instances is currently impractical. However, due to their strong pattern recognition capabilities, fine-tuned LLMs can provide satisfactory feasible solutions that serve as upper bounds for minimization problems.

We aim to explore the potential of LLMs in this context. If LLMs can provide satisfactory feasible solutions during the model-building process, or if traditional solvers like Gurobi face challenges in finding feasible solutions for large-scale instances, these LLM-generated solutions could serve as valuable upper bounds to accelerate the pruning process. By comparing the solutions generated by LLMs with the top three feasible solutions produced by traditional mathematical solvers, we can potentially leverage LLM-generated solutions to enhance pruning strategies and improve overall solver efficiency.

\subsection{Comparision with Gurobi}
We fine-tuned the LLaMA-3.1-Instruct-8B model for the following experiment and split 10\% of the total instances in the generated dataset, which contains 12,500 MIP instances, as our test dataset.

We compare the best feasible solution obtained from the LLM using three recursive calls, where the temperature gradually decreases from 1 to 0.1 to 0.01, against the first three feasible solutions generated by GUROBI \cite{achterberg2019s}, CPLEX \cite{manual1987ibm}, and COPT \cite{ge2022cardinal} to evaluate which approach achieves a smaller gap. The gap is calculated using the formula:

\[
\text{gap} = \frac{\text{current objective value} - \text{optimal value}}{\text{current objective value}}
\]

Figure \ref{imgs:comparisionLLMandGurobi} illustrates the relationship between the scale of the MIP instance and the gap observed from LLMs, GUROBI, CPLEX, and COPT. The fine-tuned LLaMA 3.1 (8B) model is able to generate feasible solutions that are much closer to the optimal solution compared to the first three feasible solutions generated by traditional mathematical solvers.

% low-up-low temperature LLM v.s. gurobi 前3个node
\subsection{Ablation}

To demonstrate the effectiveness of the temperature adjustment strategy from higher to lower in enhancing the performance of our fine-tuned LLaMA 3.1 8B model, we conduct a comparison across several scenarios. These include cases (1) where the temperature initially rises from 0.01 to 1 and then falls back to 0.01 recursively, (2) where the temperature remains constant at 0.01 during recursive calls, (3) where a single temperature setting of 0.01 is used, and (4) where the temperature progressively rises from 0.01 to 0.1 to 1 with each 

\begin{table}[h]
    \centering
    \caption{Average Scores for Different Temperature Strategies}
    \begin{tabular}{|c|c|}
        \hline
        Strategy & Average Score \\
        \hline
        Single Temperature Call & 0.560 \\
        \hline
        Constant Temperature & 0.732 \\
        \hline
                Temperature Rise Then Fall & 0.756 \\

        \hline
                Temperature Rise & 0.813 \\

        \hline
                \textbf{Temperature Fall} & \textbf{0.840} \\

        \hline
    \end{tabular}
    \label{tab:average_scores}
\end{table}
\input{tables/TemperatureStrategy}

recursion. This comparative analysis highlights the impact of each strategy on the model's ability to generate optimal feasible solutions.

The comparison results for medium-scale MIP instances are shown in Figure \ref{imgs:ablation_experiment}. The solution quality score represents the percentage of feasible solutions generated by LLMs that have a smaller gap compared to the Gurobi solver. The average score in the test dataset is illustrated in Table \ref{tab:average_scores}. The optimal approach involves using a high temperature to encourage LLMs to explore a broader range of possibilities, which helps prevent them from getting trapped in specific nodes. Subsequently, lowering the temperature exploits and refines the better results, allowing for improved solution quality.

% low-up-low tempertature 
%constant recursively temperature 
% constant non-recursive
% up->low
% low -> up

%% file: tables/TemperatureStrategy.tex
\begin{figure}[h!]
    \centering
    \includegraphics[width=0.4\textwidth]{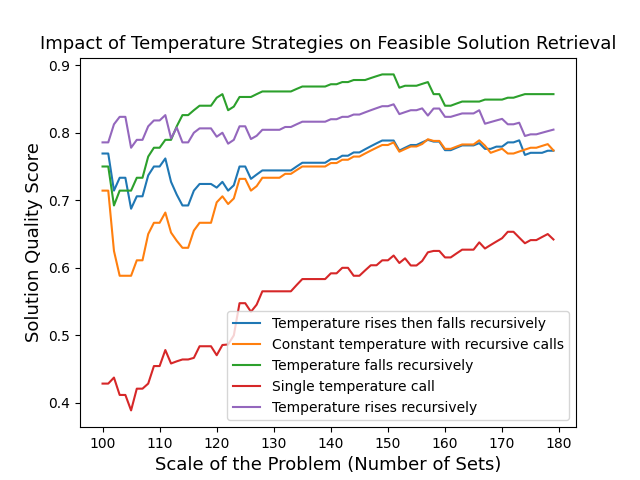} 
\caption{The solution quality score represents the percentage of feasible solutions generated by LLMs that have a smaller gap compared to the Gurobi solver, which shows a larger score means better performance.}

    \label{imgs:ablation_experiment}
\end{figure}

%% file: sections/6conclusion.tex
\section{Conclusion}
\label{sec:conclusion}

Our study evaluates the potential of LLMs to address unknown MIP models and uses the carpooling dispatch case to demonstrate how LLMs can enhance efficiency in finding better feasible solutions. This, in turn, can expedite traditional mathematical solvers' processes by pruning unnecessary nodes. We also examine the effectiveness of various temperature management strategies for fine-tuned LLama-3.1-8B in solving medium-scale MIP instances. The comparison results reveal that different temperature management approaches significantly influence the quality of feasible solutions obtained by the model. Starting with a high temperature to explore more nodes and then exploiting better solutions by gradually lowering the temperature improves the quality of feasible solutions generated by LLMs.